\documentclass[a4paper, 10pt, conference]{ieeeconf}
\IEEEoverridecommandlockouts
\usepackage{graphicx} 
\usepackage{subfigure}
\usepackage{xcolor}
\usepackage{multirow}
\usepackage{siunitx}
\usepackage{booktabs}
\usepackage{tabularx}
\usepackage{hyperref}
\usepackage{fullwidth}

\title{PHI Scrubber: A Deep Learning Approach}

\author{
Abhai Kollara Dilip\textsuperscript{*} \hspace{1em} Kamal Raj K\textsuperscript{*} \hspace{1em} Malaikannan Sankarasubbu\\
Saama Technologies AI Research Lab \\
Chennai, India\\
{\tt\small \{a.kollara, kamal.raj, malaikannan.sankarasubbu\}@saama.com}
}

\begin{document}
\maketitle
\thispagestyle{empty}
\pagestyle{empty}
\begin{abstract}
\renewcommand{\thefootnote}{\roman{footnote}}
\footnotetext{*Both authors had equal contribution}
Confidentiality of patient information is an essential part of Electronic Health Record System. Patient information, if exposed, can cause a serious damage to the privacy of individuals receiving healthcare. Hence it is important to remove such details from physician notes. A system is proposed which consists of a deep learning model where a de-convolutional neural network and bi-directional LSTM-CNN is used along with regular expressions to recognize and eliminate the individually identifiable information. This information is then removed from a medical practitioner's data which further allows the fair usage of such information among researchers and in clinical trials.
\end{abstract}

\section{Introduction}
Patient identification for clinical trials is one of the major challenges in pharma industry since major percentage of clinical trials fail due to patient enrollment issues. The structured EHR data that are used to identify potential candidates for trials is designed to support billing processes with Health Insurance companies and may not include all the necessary information for identification. Doctor notes on the other hand contain a wealth of information but remain inaccessible due to the presence of protected health information that must remain confidential. The HIPAA act of 1996 sets forth national standards for the transaction of electronic healthcare information. As a direct result of this, any document containing information that can be used to trace the patient has to be treated as protected health information (PHI). Such documents must remain confidential unless there is a clear consent from the patient involved. HIPAA specifies a set of 18 identifiers whose presence make a document PHI (Refer methods section). De-identifying such documents involve removing the identifiers. The notes however are unstructured data and thus requires a system that can learn patterns in the language.

\section{Data}
Authentic protected health information is not available for research (in large amounts) due to the very problem we are trying to solve. Since deep learning models require a large amount of data to perform well, we are forced to use a substitute dataset. The task at hand is an entity recognition problem and thus we use other datasets that are widely used for the problem.

Ontonotes corpus \cite{ontonotes} is a large manually annotated corpus containing text from a variety of genres (news, talk shows, newsgroups, conversational phone calls) in 3 languages (English, Chinese and Arabic). The dataset contains several layers of annotations for various natural language processing tasks. For the purpose of our experiments, we use the entity names layer of English language to train the models.

The dataset contains sentences and the corresponding entity label of each token in the sentence. The sentences are segmented into tokens using the Penn Treebank tokenizer. The tokenization splits the sentences based on whitespace as well as punctuation (eg: Robert's friend cannot be there -> "Robert","'s","friend", "can","not","be", "there").

The named entities in the dataset are classified into 18 possible categories which are labelled using BILOU labelling scheme. The BILOU scheme labels each word as either of Beginning (\textbf{B}), Inside (\textbf{I}) or Last (\textbf{L}) of an entity if the entity is multi-word. Unit length entities are marked as \textbf{U} while a non-entity is labelled as \textbf{O}.

\begin{table}[h!]
    \centering
    \begin{tabular}{|c|l|}
    \hline
         PERSON & Person \\
         NORP & Nationalities or religious or political groups \\
         FACILITY & Buildings, airports, highways, bridges, etc \\
         ORGANIZATION & Companies, agencies, institutions, etc. \\
         GPE & Countries, cities, states \\
         LOCATION & Non-GPE locations, mountain ranges, bodies of water\\
         PRODUCT & Vehicles, weapons, foods, etc. (Not services)\\
         EVENT & Named hurricanes, battles, wars, sports events, etc.\\
         WORK OF ART & Titles of books, songs, etc.\\
         LAW & Named documents made into laws\\
         LANGUAGE & Any named language \\
         DATE & Absolute or relative dates or periods \\
         TIME & Times smaller than a day \\
         PERCENT & Percentage \\
         MONEY & Monetary values, including unit \\
         QUANTITY & Measurements, as of weight or distance \\
         ORDINAL & “first”, “second” etc \\
         CARDINAL & Numerals that do not fall under another type \\
    \hline
    \end{tabular}
    \caption{Entity types in Ontonotes dataset}
    \label{Table 1}
\end{table}
\section{Methods}
A PHI scrubber should remove the following identifiers from a given document.
\begin{figure}
    \centering
    \includegraphics[width=0.5\textwidth]{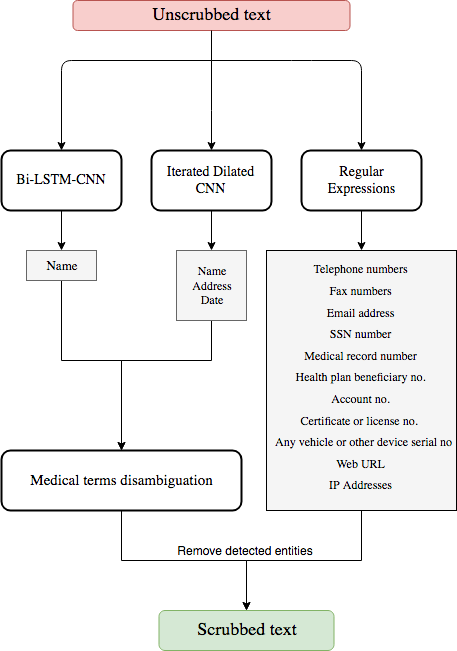}
    \caption{Scrubber architecture}
    \label{fig:Architecture}
\end{figure}

\begin{enumerate}
    \item Name
    \item Address (all geographic subdivisions smaller than state, including street address, city county, and zip code)
    \item All elements (except years) of dates related to an individual (including birthdate, admission date, discharge date, date of death, and exact age if over 89)
    \item Telephone numbers
    \item Fax number
    \item Email address
    \item Social Security Number
    \item Medical record number
    \item Health plan beneficiary number
    \item Account number
    \item Certificate or license number
    \item Any vehicle or other device serial number
    \item Device identifiers and serial numbers
    \item Web URL
    \item Internet Protocol (IP) Address
    \item Finger or voice print
    \item Photographic image - Photographic images are not limited to images of the face.
    \item Any other characteristic that could uniquely identify the individual
\end{enumerate}

For text data we ignore 16, 17 and 18.
The dataset we chose contains 1, 2 and 3 thus allowing us to use deep learning models to identify them. The models can capture variations of the entities that cannot be captured using traditional methods. For the rest of the identifiers regular expressions are sufficient since all of them follow a predictable and limited set of pattern.

\subsection*{Deep learning models}
A combination of two deep learning models are applied here - A bidirectional variation of long short-term memory combined with character level convolutional neural networks \cite{DBLP:journals/corr/ChiuN15} and an iterated dilated convolutions model (ID-CNN) \cite{strubell}

Since recurrent neural networks are known for its performance in sequence labeling tasks, initially we applied the bi-LSTM CNN model for this task. But due to the extremely high number of false positives, it produced we used it solely for identifying names. An iterated dilated convolutions model was then added to the model to improve its performance. This model was chosen for its state of the art results in named entity recognition. Both models are trained separately to identify all the entity types available in the Ontonotes dataset. But for inference, Bi-LSTM detects "NAME" entity while ID-CNN model captures "NAME", "ADDRESS" and "DATE" entities.

\subsection*{Bidirectional LSTM-CNN}
The Bi-LSTM-CNN model has the ability to capture context from the entire sequence as well as incorporate character level features. For the model, we create 3 different vector representations for each word in a sentence
\begin{itemize}
    \item Word embedding
    \item Character level embedding
    \item Additional word features
\end{itemize}

Word embeddings are obtained using a lookup table. The pre-trained Glove embeddings \cite{pennington2014glove} are used for this purpose.

For character level embeddings, we represent each character in a word by a vector, which are then concatenated. A 1D convolution is applied over each word and is followed by max pooling to obtain character level representations.(Fig.3)

Finally, we use additional word features by categorizing each word into one of the following classes, which are then mapped to a random vector.

\begin{itemize}
    \item Numeric
    \item All lower case
    \item All upper case
    \item Initial upper case
    \item Mostly numeric
    \item Contains digit
    \item Other
\end{itemize}

The three different representations are then concatenated and fed into a stacked bidirectional LSTM\cite{Hochreiter:1997:LSM:1246443.1246450}\cite{graves2005framewise} layer. The outputs at each timestep are passed through a softmax layer to get the score for each word. Each word is classified into one of 5 varieties (we use BILOU labeling scheme) of the 18 different categories available in the Ontonotes dataset. Even though we require only 3 of the 18 entity types, we train the model using all entity types.

\begin{figure}
    \centering
    \includegraphics[width=0.5\textwidth]{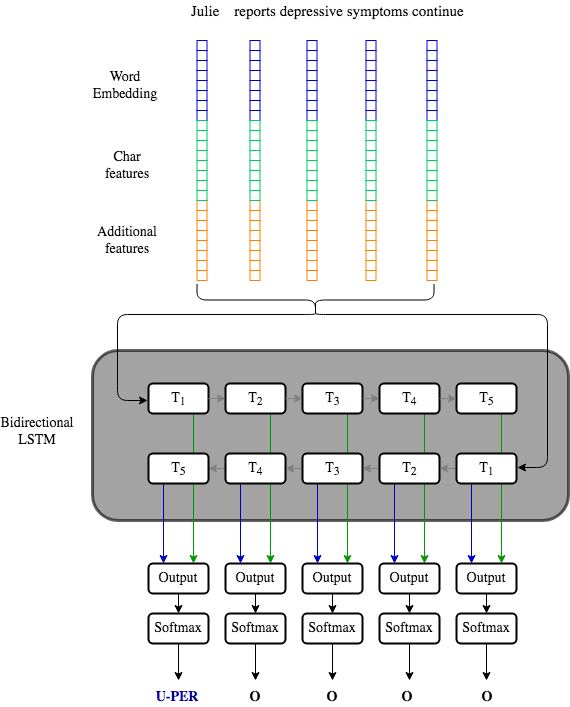}
    \caption{Bidirectional LSTM}
    \label{fig:Bi-LSTM}
\end{figure}
We used bidirectional LSTM-CNN model since the CNNs are able to capture character level features. The model combines character level features with word-level features thus making it tolerant to minute spelling variations.

\begin{table}[h!]
    \centering
    \begin{tabular}{|c|c|}
    \hline
         Word embedding size & 300 \\
         Character embedding size & 95 \\
         LSTM cell size & 200 \\
         LSTM Layers & 2\\
         Learning rate & 0.001\\
         Optimizer & Nadam\\
         Epochs & 65\\
    \hline
    \end{tabular}
    \caption{Training parameters for Bi-LSTM-CNN network}
    \label{Table 2}
\end{table}

\begin{figure}
    \centering
    \includegraphics[width=0.4\textwidth]{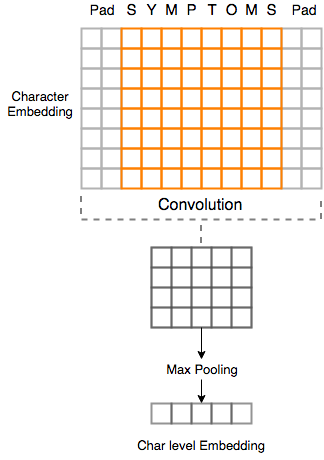}
    \caption{Character CNN}
    \label{fig:char cnn}
\end{figure}

\subsection*{Iterated Dilated Convolution}
Here, we replicate the model created by Strubell et al.\cite{strubell}. Similar to the previous model, this network takes in a sequence of words and outputs for each word a probability distribution over all possible labels. Besides the superior results, the use of convolutions allows us to exploit parallel computation to the maximum.

The ability of recurrent neural networks to capture temporal information from the entire sequence of its inputs has made it the workhorse of NLP tasks in deep learning. However, due to the nature of LSTM computations,  they are difficult to run parallelly. Convolutional networks on the other hand can easily be run in parallel but have fixed contexts ie they fail to incorporate the entire sequence as its context. Dilated convolutions provide a workaround for this obstacle\cite{dilated_convs}. Stacked dilated CNNs can easily incorporate global information from a whole sentence or document. They allow the effective input width to grow exponentially with the depth of the network. Dilated CNNs operate on similar principles of convolutional networks except that the dilated window skips over every dilation width 'd' inputs (See Fig 3.).

Similar to the Bi-LSTM-CNN model each word has multiple representations. A pretrained word embedding and additional word feature embedding. Word embeddings are obtained from pre-trained Lample embeddings\cite{lample}. The additional word features are similar to those used in Bi-LSTM CNN model. However, the following 4 categories are used instead
\begin{itemize}
    \item All upper case
    \item Initial upper case
    \item Camel case
    \item Other
\end{itemize}
The two different representation are concatenated and fed into iterated dilated CNN layers. The ID-CNN architecture repeatedly applies the same block of dilated convolution to the input representations. The transition parameter and score of each token is passed through a Viterbi decoding to get the score for each word. Similar to the Bi-LSTM-CNN model, we use a BILOU labeling scheme here.

\begin{figure}
    \includegraphics[width=0.5\textwidth]{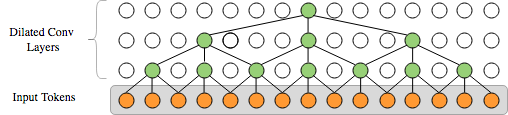}
    \caption{Iterated Dilated Convolution}
    \label{fig:dilated cnn}
\end{figure}

\begin{table}[h!]
    \centering
    \begin{tabular}{|c|c|}
    \hline
         Word embedding size & 100 \\
         ID-CNN layers & 3 \\
         ID-CNN dilation & [1,2,1]\\
         ID-CNN width & 3\\
         ID-CNN filters & 400\\
         ID-CNN blocks & 1\\
         Learning rate & 0.0001\\
         Optimizer & adam\\
         Epochs & 100\\
    \hline
    \end{tabular}
    \caption{Training parameters for ID-CNN network}
    \label{Table 3}
\end{table}

\subsection*{Regular expressions}
Regular expressions are used to detect all the below-given identifiers. We observed regular expression were sufficient to detect the following identifiers.
\begin{itemize}
    \item Telephone numbers
    \item Fax numbers
    \item Email address
    \item SSN number
    \item Medical record number
    \item Health plan beneficiary number
    \item Account number
    \item Certificate or license number
    \item Any vehicle or other device serial number
    \item Web URL
    \item Internet Protocol (IP) Address
\end{itemize}

\subsection*{Medical terms disambiguation}
Due to the absence of medical terms in Ontonotes dataset several such terms are misclassified by the deep learning models. The presence of names in disease and drug names (eg: Parkinson's disease) are also a cause for misclassification. The sentence structure at times causes body part names to be classified as a location. To circumvent this problem, we use a separate disambiguation module for the outputs of the named entity recognition module. The "PERSON", "LOCATION" and "DATE" entities that are detected by the deep learning models are passed into this disambiguation module. Here, a fuzzy search is done over a dictionary of medical terms. For non-abbreviated words, we consider the Levenshtein distance for matching. If the word has a Levenshtein distance less than 2 for any word in the dictionary we consider this as a match. For abbreviated words only exact matches are considered. Besides this, we also check if a word ends with drug name stem (eg.-dralazine, -pristone etc) to further check for drug names. Once all the matches have been found, their labels are removed thus preventing the entities from being removed.

\begin{figure}
    \includegraphics[scale=0.5]{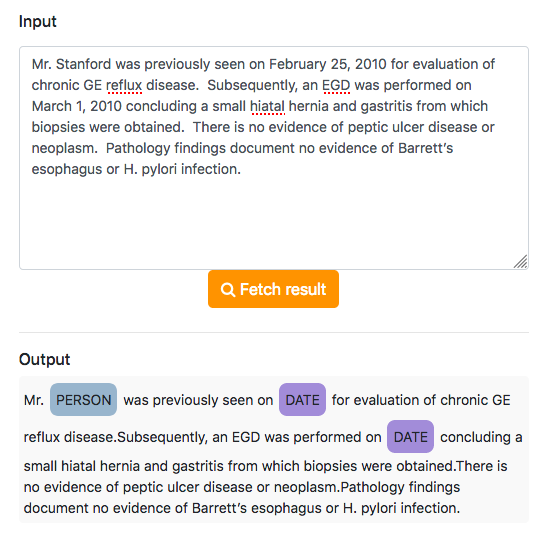}
    \caption{PHI De-Identifier}
    \label{fig:PHI De-Identifier}
\end{figure}

\section{Training}
We trained the BiLSTM-CNN and ID-CNN model for 100 epochs using Adam optimizer. Both models are trained independently on the Ontonotes dataset to identify all types of entities in the dataset. While training only the best weights are saved. The training samples are batched into sentences of equal length. Parameters for training the models are given in tables II and III. A single Nvidia Geforce GTX 1080 took about 6 hours to complete the training for BiLSTM-CNN while the ID-CNN model took 1.5 hours.

\section{Results}
Evaluation of the models was done on the Ontonotes dataset. The ID-CNN model achieves a segmented micro F1 score of 86.84 while Bi-LSTM model achieves 86.5. The models are evaluated on the performance on Ontonotes dataset. Even though both models produce near similar results, the ID-CNN model is significantly faster during training.

\section{Conclusion}
We present a deep learning based approach for the removal of PHI from text documents. Currently, we use a regular expression to assist the removal of identifiers. In future, we hope a single unified deep learning model can identify all identifiers with higher accuracy.

\section*{Acknowledgment}
The authors would like to thank Dr.Anand Dubey and V Archana for reviewing the manuscript and for providing their technical inputs.

\end{document}